\documentclass[11pt,letterpaper]{article}

\usepackage[T1]{fontenc}
\usepackage[utf8]{inputenc}
\usepackage{lmodern}
\usepackage[english]{babel}
\usepackage{algpseudocode}

\usepackage{graphicx}%
\usepackage{multirow}%
\usepackage{amsmath,amssymb,amsfonts}%
\usepackage{amsthm}%
\usepackage{mathrsfs}%
\usepackage[title]{appendix}%
\usepackage{xcolor}%
\usepackage{textcomp}%
\usepackage{manyfoot}%
\usepackage{booktabs}%
\usepackage{algorithm}%
\usepackage{algorithmicx}%
\usepackage{algpseudocode}%
\usepackage{listings}%
\usepackage{graphicx}
\usepackage{placeins}

\usepackage{amsmath,amssymb,amsthm}

\usepackage{graphicx}
\usepackage{booktabs}
\usepackage{hyperref}

\usepackage[margin=2.5cm]{geometry}

\DeclareMathOperator*{\argmax}{arg\,max}

\title{Semi-Linear Gaussian Bayesian Networks: Enhancing Flexibility in Node Dependencies via penalized B-Splines}
\author{Aitor Oviedo-Madrid\textsuperscript{1,2,*}, Mabel Morales-Otero\textsuperscript{1,2}, Jose Gonzalez-Gomariz\textsuperscript{1,2},\\Rubén Armañanzas\textsuperscript{1,2}}

\begin{document}

\maketitle

\begin{center}
\small
\textsuperscript{1} Institute of Data Science and Artificial Intelligence (DATAI), University of Navarra, Pamplona, Spain\\
\textsuperscript{2} TECNUN School of Engineering, University of Navarra, San Sebastián, Spain\\
\textsuperscript{*} Corresponding author: \texttt{aoviedomadr@unav.es}
\end{center}

\begin{abstract}
Bayesian networks for continuous variables often rely on restrictive assumptions: linear Gaussian models lack flexibility to capture nonlinear dependencies, while non-parametric approaches provide greater flexibility at a high computational cost. This creates a gap for modeling scenarios where variables are approximately Gaussian but exhibit nonlinear relationships. We introduce Semi-Linear Gaussian Bayesian Networks (SLGBN), a novel class of continuous Bayesian networks designed to bridge this gap, preserving Gaussian conditional distributions while introducing nonlinear conditional mean functions. The proposed framework provides a unified probabilistic framework combining linear Gaussian and nonlinear Gaussian conditional models within the same Bayesian network Nonlinear conditional mean functions are estimated using generalized additive models (GAMs) with penalized B-splines. We present parameter estimation procedures for both types of nodes, as well as a greedy hill-climbing algorithm for learning the network structure under the SLGBN framework. To assess its performance, we compare SLGBN with other continuous Bayesian network models, including semiparametric Bayesian networks (SPBN) and Linear Gaussian Bayesian networks (LGBN). Experimental results evaluate both goodness of fit and the ability to recover the underlying network structure. The results show that, in the presence of nonlinear Gaussian relationships, SLGBN achieves better fit and improved structure recovery compared to alternative models, particularly when combined with the Extended Bayesian Information Criterion (EBIC*) score. Furthermore, by retaining Gaussian conditional distributions, SLGBN offers computational advantages over fully semiparametric approaches.
\end{abstract}


\section{Introduction}\label{sec:intro}

Bayesian networks are probabilistic graphical models that provide a compact representation of joint probability distributions. Their graphical structure encodes conditional independence relationships among variables, making them a powerful and interpretable tool for modeling complex multivariate systems~\cite{koller2009probabilistic}. Their ability to combine probabilistic reasoning with interpretable graphical representations has led to their widespread adoption in data mining and decision-support applications~\cite{heckerman1997}. Despite these advantages, Bayesian networks also present important limitations, particularly when dealing with continuous variables.

Inference and learning in Bayesian networks with continuous variables is considerably more challenging than in the discrete case. Exact inference typically requires restrictive assumptions on the form of the conditional distributions, and structure learning becomes computationally demanding due to the need to estimate complex likelihood functions. As a consequence, continuous variables are commonly discretized to simplify learning and inference. While discretization can be effective in some settings, it may lead to information loss and can alter relevant features of the original data distribution~\cite{NOJAVANA201764}.

To overcome these limitations, a growing body of research has focused on extending Bayesian networks to model continuous variables. Among continuous Bayesian networks, Linear Gaussian Bayesian Networks (LGBN) constitute one of the most widely used approaches. In this framework, each variable is modeled as a Gaussian distribution whose mean is a linear function of its parents. This assumption greatly simplifies inference and parameter estimation, making LGBN computationally attractive. Although linear Gaussian models often provide a reasonable approximation to real world data, they rely on strong assumptions of linearity and normality that are frequently violated in practice.

More flexible alternatives have been recently proposed. In particular, Semiparametric Bayesian Networks (SPBN) allow conditional distributions to vary between linear Gaussian forms and nonparametric estimates~\cite{ATIENZA2022564}, commonly obtained via kernel density estimation (KDE). Since their introduction, SPBNs have motivated several extensions, including hybrid formulations~\cite{atienza2022hybrid} or transfer learning approaches~\cite{SOJO2026116077}, highlighting their current interest. These models have shown improved performance in capturing nonlinear dependencies and deviations from Gaussianity. However, this increased flexibility comes at the cost of significantly higher computational complexity and more challenging inference procedures.

Despite these advances, an important modeling gap remains. There are important scenarios where variables are approximately Gaussian but the relationships among them are nonlinear. In such scenarios, LGBNs are often too restrictive because of their linear conditional models, whereas fully semiparametric approaches may introduce unnecessary complexity. This motivates the need for models that preserve the computational advantages of Gaussian Bayesian networks while allowing nonlinear conditional relationships. Recent work has explored this direction by incorporating nonlinear regression models, such as spline-based regressors, into Bayesian network learning procedures~\cite{sharma2022scalable}. However, these approaches primarily employ nonlinear models as components of structure learning algorithms rather than defining a unified probabilistic framework for continuous Bayesian networks.

In this paper, we introduce Semi-Linear Gaussian Bayesian Networks (SLGBN), a novel class of continuous Bayesian networks that bridges the gap between linear Gaussian and fully semiparametric models. SLGBN provides a unified probabilistic framework that combines linear Gaussian and nonlinear Gaussian conditional models within the same network while preserving Gaussian conditional distributions. Nonlinear conditional means are modeled using Generalized Additive Models (GAMs) with penalized B-splines, providing greater flexibility while retaining computational advantages over fully semiparametric models. To make the proposed framework applicable in practice, we derive parameter estimation procedures for both linear and nonlinear Gaussian nodes and develop a greedy hill-climbing algorithm for simultaneously learning the network structure and node types. Finally, we evaluate SLGBN on synthetic Bayesian networks and compare its performance with LGBN and SPBN models in terms of both goodness of fit and structure recovery. The main contributions of this work are summarized as follows:

\begin{itemize}
\item We introduce SLGBN, a novel class of continuous Bayesian networks that bridges the gap between linear Gaussian and fully semiparametric models. 
\item We define a unified probabilistic framework combining linear and nonlinear Gaussian conditional models while preserving Gaussian conditional distributions. 
\item We derive parameter estimation procedures and develop a greedy hill-climbing algorithm for jointly learning network structure and node types. 
\item We experimentally show that, for approximately Gaussian variables exhibiting nonlinear dependencies, SLGBN achieves improved goodness of fit and structure recovery while maintaining a favorable balance between modeling flexibility and computational complexity. 
\end{itemize} 

\section{Background}\label{sec:background}

This section introduces Bayesian networks, along with linear Gaussian Bayesian networks (LGBN) and semiparametric Bayesian networks (SPBN), two continuous Bayesian network models, which serve as reference models for comparison with SLGBN. It then presents generalized additive models (GAM) and penalized B-splines, which form the basis for estimating nonlinear nodes in the SLGBN framework.

\subsection{Bayesian Networks}

\noindent
A Bayesian Network $\mathcal{B}=(\mathcal{G}, \theta)$ consists of a Directed Acyclic Graph (DAG) $G=(V,E)$ with node set $V = \{1, \cdots, p\}$ and edge set $E$. A set of parameters $\theta$ defines the conditional probability distributions between them. For a set of variables $\mathbf{X} = \{X_{1}, ..., X_{p}\}$, a Bayesian Network encodes its joint probability $P(\mathbf{X})$. The distribution is factorized as $P(\mathbf{X}) = \prod_{j=1}^{p} P(X_{j}\mid\mathbf{Y}_j)$, where $\mathbf{Y}_j$ denotes the parent variables of $X_j$ in $\mathcal{G}$.

\subsection{Linear Gaussian Bayesian Networks}

\noindent
A Linear Gaussian Bayesian Network (LGBN) is a specific kind of continuous Bayesian network in which all the CPD are linear Gaussians. Let $X_j$ be a linear Gaussian of its parents $\mathbf{Y}_j$.

\begin{equation}
    P(X_j|\mathbf{Y}_j) = \mathcal N\left(\beta_{j,0}+\boldsymbol{\beta}_{j}^{\top}\mathbf{Y}_{j}, \sigma_{j}^{2}\right).
    \label{eq:LG_densities}
\end{equation}

\noindent
Assume that $\mathbf{Y}_{j}$ is jointly Gaussian with distribution $\mathcal N\left(\boldsymbol{\mu}_{\mathbf{Y}_j},\boldsymbol{\Sigma}_{\mathbf{Y}_j}\right)$. Then the distribution of $X_j$ is Gaussian with 

\[
\mathbb E[X_j]
=
\beta_{j,0}
+
\boldsymbol{\beta}_j^{\top}
\boldsymbol{\mu}_{Y_j},
\]

\[
\mathrm{Var}(X_j)
=
\sigma_j^2
+
\boldsymbol{\beta}_j^{\top}
\boldsymbol{\Sigma}_{Y_j}
\boldsymbol{\beta}_j,
\]

\noindent
and the joint distribution of $(X_j,\mathbf Y_j)$ is a multivariate Gaussian. Therefore, if $\mathcal B$ is a linear Gaussian Bayesian network, then it defines a joint distribution that is jointly Gaussian. Multivariate Gaussian distributions are closed under marginalization and conditioning, which guarantees the existence of closed-form expressions for all conditional and marginal distributions. This property is particularly advantageous for exact inference and structure learning in Bayesian networks.

\subsection{Semiparametric Bayesian Networks}

\noindent
Semiparametric Bayesian Networks (SPBN)~\cite{ATIENZA2022564} extend traditional Bayesian networks by allowing a mixture of parametric and non-parametric conditional probability distributions (CPD). 

For parametric nodes, the conditional distribution follows a linear Gaussian model as in LGBN. For non-parametric nodes, the conditional density is estimated using Conditional Kernel Density Estimation (CKDE), defined as:

\begin{equation}
\hat{f}(X_j \mid \mathbf{Y}_j)
=
\frac{\hat{f}(X_j, \mathbf{Y}_j)}{\hat{f}(\mathbf{Y}_j)},
\end{equation}

\noindent
where $\hat{f}(X_j, \mathbf{Y}_j)$ is estimated using kernel density estimation. This flexible formulation allows SPBN to capture both linear Gaussian dependencies and complex nonlinear relationships. However, the use of non-parametric density estimation significantly increases computational complexity and limits scalability in structure learning tasks.

\subsection{Generalized Additive Models}

\noindent
Generalized Linear Models (GLM) extend the linear regression framework allowing for response variables from the exponential family. For a response variable $X_j$ with parent vector $\mathbf Y_j$, a GLM assumes a linear predictor 

\begin{equation}
    \eta_j = \beta_{j,0} + \boldsymbol{\beta}_{j}^{\top}\mathbf{Y}_j, 
\end{equation}

\noindent
together with a link function $g(\mu_j) = \eta_j$ and a variance function $\mathrm{Var}(X_j) = \phi V(\mu_j)$. Generalized Additive Models (GAM)~\cite{hastie_gam_1986} are extensions of GLM where the relation between $\eta$ and the predictors is not considered linear. Instead, a smooth function $f(\cdot)$ is used in place of the linear functions 

\begin{equation}
\eta_j = \beta_{j,0} + \sum_{k=1}^{d_j} f_{jk}(Y_{jk}),
\end{equation}

\noindent
where $d_j = |\mathbf Y_j|$ is the number of parents of $X_j$ 
and each $f_{jk}(\cdot)$ is a smooth function. This framework allows us to convert the linear dependencies of LGBN to nonlinear dependencies conditionally Gaussian distributions.

\subsection{B-Splines}
\label{subsec:bsplines}

\noindent
Splines are functions constructed from polynomial pieces joined at certain values of $x$, the knots. Let $\boldsymbol{\xi} = (\xi_0, \xi_1, \dots, \xi_t)$ be a non-decreasing knot sequence. Any spline in the spline space of order $m$ associated with $\boldsymbol{\xi}$ can be expressed as a linear combination of B-spline basis functions of degree $p = m - 1$~\cite{deboor1978apractical}.

\begin{equation}
s(x)
=
\sum_{\ell=1}^{L}
\beta_{\ell} B_{\ell,p}(x),
\end{equation}

\noindent
where $L = t - p - 1$ is the number of basis functions and 
$B_{\ell,p}(x)$ are the B-spline basis functions of degree $p$. B-splines are linear in their coefficients $\beta_{\ell}$. A B-spline can be defined as the smoothing function used by the GAM. For an $X_j$ node

\begin{equation}
\eta_j = \beta_{0} + \sum_{k=1}^{d_j} s_k(x_k).
\end{equation} 

\noindent
Using B-splines is a powerful approach for approximating functions but they tend to overfit data, especially when the number of knots is large~\cite{eilers1996flexible}. For this reason, spline coefficients are typically estimated using a penalized criterion that controls smoothness through a penalty on the coefficients.

\section{Semi-Linear Gaussian Bayesian Networks}\label{sec:SLGBN}

We now introduce the proposed Semi-Linear Gaussian Bayesian Networks (SLGBN). Learning in Bayesian networks generally involves two main tasks: estimating the local model parameters and recovering the graph structure that defines the dependency relationships among variables. In the following sections, we formally define the SLGBN framework and present the corresponding methodology for parameter estimation and structure learning, including a greedy hill-climbing algorithm adapted to jointly learn both graph structure and node-type assignments.

\subsection{Definition}

\noindent
An SLGBN is a continuous Bayesian Network $\mathcal B$ defined as a DAG $\mathcal{G}=(V,E)$ where nodes are divided into two different types $\mathbf{V} = \{\mathbf{X}_{L}, \mathbf{X}_{N}\}$: linear Gaussian nodes $\left(\mathbf{X}_{L}\right)$ and nonlinear Gaussian nodes $\left(\mathbf{X}_{N}\right)$. While linear Gaussian nodes are modeled using the well-known linear Gaussian densities (see Equation~\ref{eq:LG_densities}), the conditional mean of the nonlinear Gaussian nodes is modeled using a GAM with penalized B-splines, assuming an additive structure over the parent variables. The linearity/nonlinearity is a local property, defined with respect to the conditional dependence on each node's direct parents, rather than at the level of global dependencies in the DAG. The model reduces to a Linear Gaussian Bayesian network when all nodes $\mathbf{V}$ are modeled as linear Gaussians.

\subsection{Parameter Estimation}

\noindent
Let $\mathcal G$ be a fixed structure. For each node in the network, the parameters $\boldsymbol{\theta}_j$ are estimated from data. One common approach is to estimate the parameters that maximize the probability of the observed dataset. Let $\mathcal{D} = \{X_1, \cdots, X_p\}$ be a dataset with samples $X_{j} = \{x_1^{(j)}, \cdots, x_n^{(j)}\}$. The probability of a set of observations from node $X_j$ with a set of parents $\mathbf{Y}_j$ and a set of parameters $\boldsymbol{\theta}_j$ is simply the product of the assigned density to each data point. This is the likelihood of the node.

\begin{equation}     
\mathcal{L}\left(X_{j}\mid\boldsymbol{\theta}_{j},\mathcal{G}\right) = \prod_{i=1}^{n}f(x_i\mid\boldsymbol{\theta}_j, \mathbf{Y}_j) 
\end{equation}

\noindent
The likelihood of the complete network can be obtained by multiplying the individual likelihoods of the nodes.

\begin{equation}     
    \mathcal{L}\left(\mathcal{D}\mid\boldsymbol{\theta}_{\mathcal{G}},\mathcal{G}\right) = \prod_{j=1}^{p}\mathcal{L}\left(X_{j}\mid\boldsymbol{\theta}_{j},\mathcal{G}\right),
    \label{eq:likelihood}
\end{equation}

\noindent
and given that $\theta_j \cap \theta_k = \emptyset$ for all $j \neq k$, individually optimizing $\theta_j$ for each $j \in p$ is equivalent to maximizing $\mathcal{L}\left(\mathcal{D}\mid\boldsymbol{\theta}_{\mathcal{G}},\mathcal{G}\right)$. Usually this equation is converted to log-likelihood, as this can be expressed as a summation of terms instead of a product.

\begin{equation}     
\mathcal{LL}\left(\mathcal{D}\mid\boldsymbol{\theta}_{\mathcal{G}},\mathcal{G}\right) = \sum_{j=1}^{p}\sum_{i=1}^{n}\log{f}\left(X_{j}^{(i)}\mid\boldsymbol{\theta}_{j},\mathcal{G}\right)
\end{equation}

\noindent
It is common to estimate the parameters by maximizing the log-likelihood of the dataset. In SLGBN, nodes can be either linear or nonlinear, and may or may not have parents. 

\begin{itemize}
    \item \textbf{Nodes with no parents}.
\end{itemize}

\noindent
The probability density of every node with no parents is defined as a univariate Gaussian. For a $X_{j}$ node with no parents:

\begin{equation}
P(X_{j}) \sim N(\mu_{j}, \sigma^{2}_{j})
\end{equation}

\noindent
where parameters are given by sample mean and variance. Under the normality assumption, these are the maximum likelihood parameters~\cite{hogg2005mathematicalstats}. Let $X_{j}$ be a node with a set of parents $\mathbf{Y}_j$. In this case, the conditional mean of $X_{j}$ is modeled as a function of its parents $\mu_{j|\mathbf{Y}_j} = \mu(\mathbf{Y}_j)$.

\begin{equation}
P(X_{j}\mid\mathbf{Y}_j) \sim N(\mu(\mathbf{Y}_j), \sigma_j^{2})
\end{equation}

\noindent
This function is determined from the type of dependency between the node and its parents, which may be linear or nonlinear.

\begin{itemize}
    \item \textbf{Linear Gaussians}.
\end{itemize}

\noindent
For a linear node $X_j$ with parent variables $\mathbf{Y}_j$, we assume a Gaussian conditional distribution with linear mean:

\begin{equation}
    P(X_j \mid \mathbf{Y}_j) \sim \mathcal{N}(\mu_j(\mathbf{Y}_j), \sigma_j^2), \quad 
    \mu_j(\mathbf{Y}_j) = \beta_{0,j} + \boldsymbol{\beta}_j^\top \mathbf{Y}_j.
\end{equation}

\noindent
In matrix notation, the model is:

\begin{equation}
    \mathbf{X} = \mathbf{Z} \boldsymbol{\theta} + \boldsymbol{\varepsilon}, \quad 
    \boldsymbol{\varepsilon} \sim \mathcal{N}(0, \sigma^2 \mathbf{I}_n),
\end{equation}

\noindent
where $\mathbf{Z}$ is the design matrix including a column of ones for the intercept, and $\boldsymbol{\theta}$ collects all coefficients. Under Gaussian assumption, the ordinary least squares (OLS) estimator of $\boldsymbol{\theta}$ is equivalent to the MLE estimator~\cite{hogg2005mathematicalstats}.

\begin{equation}
    \hat{\boldsymbol{\theta}} = (\mathbf{Z}^\top \mathbf{Z})^{-1} \mathbf{Z}^\top \mathbf{X}, \quad 
    \hat{\boldsymbol{\varepsilon}} = \mathbf{X} - \mathbf{Z} \hat{\boldsymbol{\theta}},
\end{equation}

\noindent
The variance is estimated as

\begin{equation}
    \label{eq:mle_variance}
    \hat{\sigma}^2 = \frac{\hat{\boldsymbol{\varepsilon}}^\top \hat{\boldsymbol{\varepsilon}}}{n}.
\end{equation}

\begin{itemize}
    \item \textbf{Nonlinear Gaussians.}
\end{itemize}

\noindent
For a nonlinear node $X_j$ with parents $\mathbf{Y}_j$, the conditional mean is modeled as an additive function of the parents using cubic B-splines:

\begin{equation}
    P(X_j \mid \mathbf{Y}_j) \sim \mathcal{N}(\mu_j(\mathbf{Y}_j), \sigma_j^2), \quad
    \mu_j(\mathbf{Y}_j) = \beta_{0,j} + \sum_{k=1}^{d_{j}} s_{jk}(y_{jk}).
\end{equation}

\noindent
Each smooth function $s(\cdot)$ is expressed as a linear combination of cubic B-spline bases. The number of bases is fixed a priori and defines the number of knots and the maximal flexibility of the spline. In matrix form:

\begin{equation}
    \mathbf{X} = \mathbf{Z} \boldsymbol{\theta} + \boldsymbol{\varepsilon}, \quad 
    \boldsymbol{\varepsilon} \sim \mathcal{N}(0, \sigma^2 \mathbf{I}_n),
\end{equation}

\noindent
where $\mathbf{Z}$ contains the evaluated spline basis (plus intercept) and $\boldsymbol{\theta}$ the corresponding coefficients. As commented in Section~\ref{subsec:bsplines}, the spline can tend to overfit if its smoothness is not controlled. For this reason, a penalization over the smoothness of the splines is included through a penalization parameter $\lambda$. This penalization shifts the parameter estimation problem from finding the MLE parameters to finding the parameters maximizing the penalized log-likelihood. 

\begin{equation}
\max\mathcal{LL}_{pen}\left(\mathcal{D}\mid\boldsymbol{\theta}_{\mathcal{G}},\mathcal{G}\right)\ = \max\{\mathcal{LL}\left(\mathcal{D}\mid\boldsymbol{\theta}_{\mathcal{G}},\mathcal{G}\right) - \frac{1}{2}\lambda \boldsymbol{\theta}^\top \mathbf{S} \boldsymbol{\theta}\}, 
\end{equation}

\noindent
where $\mathbf{S}$ is a penalty matrix (e.g., based on differences of coefficients or derivatives of the spline) defined over all the coefficients of the node. The parameters optimizing this penalized likelihood are obtained through a penalized least squares (PLS) approach. 

\begin{equation}
    \hat{\boldsymbol{\theta}} = (\mathbf{Z}^\top \mathbf{Z} + \lambda \mathbf{S})^{-1} \mathbf{Z}^\top \mathbf{X}, 
    \quad 
    \hat{\boldsymbol{\varepsilon}} = \mathbf{X} - \mathbf{Z} \hat{\boldsymbol{\theta}}.
    \label{eq:PLS_nonlinear}
\end{equation}

\noindent
The parameter $\lambda$ will control the smoothness of the spline, moving from a traditional OLS approach when $\lambda = 0$, to a smoother function as the value of $\lambda$ increases. The penalization only affects the estimation of the conditional mean function, and the variance is estimated using an MLE approach (see Equation~\ref{eq:mle_variance}), as in the linear case.

\subsection{Structure Learning}
\label{subsec:str_lrn}

The structure learning problem consists of identifying, from a dataset $\mathcal{D}$, the DAG $\mathcal{G}$ that best represents the dependency structure among the variables. Existing approaches can be broadly categorized into constraint-based, score-based, and hybrid methods~\cite{kitson2023survey}. In this work, we focus on the score-based paradigm. In score-based structure learning, this task is formulated as an optimization problem in which each candidate structure $\mathcal{G}$ is assigned a score $\mathcal{S}$ measuring its goodness of fit to the data. The objective is to find the structure that maximizes this score.

\begin{equation}    \max_{\mathcal{G}} \mathcal{S}\left(\mathcal{G} :\mathcal{D}\right)
\end{equation}

\noindent
Since finding the globally optimal structure under standard scoring functions is a $\mathcal{NP}$-hard problem~\cite{Chickering1996}, practical algorithms rely on heuristic search strategies. These methods explore the space of DAG through local modifications of an initial structure, selecting changes that improve the score. One common selection for the score function is the likelihood (see Equation~\ref{eq:likelihood}). Usually, this is developed by finding, from candidate structures, the set  $(\mathcal{G}, \boldsymbol{\theta})$ that maximizes the likelihood of data. This is obtained by finding first the MLE parameters for each candidate structure and then selecting the structure $\mathcal{G}$ with the maximum log-likelihood.

\begin{equation}    \max_{\mathcal{G}, \boldsymbol{\theta}} \mathcal{LL}\left(\{\mathcal{G}, \boldsymbol{\theta}\} :\mathcal{D}\right) = \max_{\mathcal{G}} \mathcal{LL}\left(\{\mathcal{G}, \boldsymbol{\hat{\theta}_{MLE}}\} :\mathcal{D}\right)
\end{equation}

\noindent
In SLGBN, the parameters of nonlinear nodes do not correspond exactly with MLE parameters but with those that maximize the penalized likelihood. Therefore, the problem shifts to finding the structure that maximizes the likelihood for the estimated parameters $\boldsymbol{\hat{\theta}}$.

\begin{equation}\label{eq:LL}\max_{\mathcal{G}} \mathcal{LL}\left(\{\mathcal{G}, \boldsymbol{\hat{\theta}}_{\mathcal{G}}\} :\mathcal{D}\right)
\end{equation}

\noindent
However, likelihood-based learning tends to overfit and typically favors overly dense graphs. To control model complexity, regularized scores such as the Bayesian Information Criterion (BIC) are commonly used:

\begin{equation}
    \max_{\mathcal{G}} \mathcal{S}_{BIC}(\mathcal{G} :\mathcal{D}) = \max_{\mathcal{G}}\left(\mathcal{LL}(\{\mathcal{G}, \boldsymbol{\hat{\theta}}_{\mathcal G}\}: \mathcal D) - \frac{k}{2}\cdot{\log{n}}\right),
\end{equation}

where $n$ denotes the sample size and $k$ the total number of free parameters. The BIC penalty is local and decomposable across nodes, so for a network with $p$ nodes, the total value for $k$ will correspond with the total sum of parameters of all the individual nodes in the network $k = \sum_{j}^{p}{k_j}$. This property allows efficient evaluation and updates during structure learning, since each node contributes independently to the total score. The total number of free parameters of each node depends on its type and its parents. Linear nodes and nodes without parents will contribute $d_j +2$ parameters each, where $d_j$ corresponds with the number of parents. For nonlinear nodes, the number of effective parameters is given by $\text{edf} + 1$, where $\text{edf}$ accounts for the complexity of the spline-based conditional mean. Following calculations in Equation~\ref{eq:PLS_nonlinear}, $\text{edf}$ is defined as

\begin{equation}
    \mathbf{H} = \mathbf{Z} (\mathbf{Z}^\top \mathbf{Z} + \lambda \mathbf{S})^{-1} \mathbf{Z}^\top, 
    \quad
    \text{edf} = \operatorname{tr}(\mathbf{H}).
    \label{eq:edf}
\end{equation}

The additional parameter corresponds to the dispersion parameter $\sigma_j^2$ of the conditional distribution. The extended BIC (EBIC)~\cite{Chen_EBIC} extends the classical BIC criterion by introducing an additional penalty on model complexity. A related formulation has been proposed for Gaussian graphical model selection, where the global penalty is defined over graph density and explicitly accounts for the size of the edge space~\cite{Foygel_EBIC}. Such additional regularization can be particularly useful in Bayesian network structure learning, where the number of candidate DAG grows super-exponentially with the number of variables. In this work, we adopt a variant of this formulation (referred to here as EBIC*), in which the local penalty is defined in terms of the number of free parameters $k$, as in the standard BIC formulation, while the global penalty depends on the total number of edges $|\mathbf{E}|$.

\begin{equation}
\max_{\mathcal{G}}\mathcal{S}_{EBIC^{*}}(\mathcal{G} :\mathcal{D}) = \max_{\mathcal{G}}\left(\mathcal{LL}(\{\mathcal{G}, \boldsymbol{\hat{\theta}}_{\mathcal G}\}: \mathcal D) - \frac{k}{2}\log n - 2\gamma |\mathbf{E}|\log p\right).
\end{equation}

This formulation allows a more flexible penalization of model complexity by assigning higher cost to nodes with richer parameterizations, which better reflects the heterogeneous nature of SLGBN. As with BIC, the likelihood and local penalty terms in EBIC* remain decomposable across nodes. Only the additional graph-structure penalty is global, depending on the total number of edges, and can still be updated efficiently during structure learning.

Overall, these two criteria represent standard choices for structure learning in Bayesian networks. In general, other scoring functions that trade off data fit and model complexity can also be used within this framework.

\subsubsection{Identifiability and Score Equivalence}
\label{subsubsec:Identifiability}

A Markov equivalence class (MEC) consists of all DAG that encode the same set of conditional independence relations over a set of variables, and is uniquely characterized by a common skeleton and identical v-structures. Within this framework, two DAG $G_1$ and $G_2$ are equivalent, denoted $G_1 \sim G_2$, if they belong to the same MEC. A scoring function $S$ is said to be score-equivalent if it assigns identical values to all DAG within the same equivalence class, that is, if $G_1 \sim G_2$ implies $S(G_1)=S(G_2)$~\cite{Chickering_2002_optimal}. This property ensures that observationally indistinguishable DAG cannot be separated by the score.

In classical LGBN, homogeneous local parameterizations imply likelihood equivalence across Markov equivalent structures. As a result, scores such as BIC and EBIC* are score-equivalent under standard assumptions for likelihood-based estimation. In particular, the additional graph-complexity term in EBIC* depends only on the number of edges, which remains invariant within a MEC. Consequently, the optimal score-based estimator is asymptotically consistent at the MEC level rather than at the level of a unique DAG~\cite{Chickering_2002_optimal, Chick_1995_Atransformational}.

In contrast, both SPBN and SLGBN introduce heterogeneous local parameterizations. Since different DAG may induce different combinations of local functional forms, score equivalence may no longer hold across Markov equivalent structures. Consequently, observationally equivalent DAG can receive different scores, and recovery of the exact generating DAG cannot in general be guaranteed from observational data alone~\cite{Hecker_1995_Learning}.

For this reason, evaluation of learned structures is primarily performed using MEC-level or skeleton-based metrics~\cite{Chickering_2002_Learning}, which better reflect the intrinsic identifiability limitations of the problem.

\subsubsection{Greedy Hill-Climbing for SLGBN.}

\noindent
We define the structure learning procedure for SLGBN as a variant of the tabu greedy hill-climbing algorithm proposed by Atienza \textit{et al.} for SPBN~\cite{ATIENZA2022564}. The corresponding pseudocode is given by Algorithm~\ref{alg:ghc_tabu}. The objective is to find the best set of edges and node-type assignments for a given dataset by maximizing a user-defined scoring function $S$, which guides the structure learning process. At each iteration, candidate structures are generated through local modifications of the current graph $\mathcal{G_\text{current}}$ and evaluated according to $S$, selecting the highest-scoring admissible move subject to an acyclicity constraint. We also enforce a maximum parent constraint ($d_{max}$) during the search. This restriction is a common assumption in score-based structure search to reduce the search space and avoid overfitting~\cite{koller2009probabilistic}. Given the structure in each iteration $G_{current}$, $\mathcal{O}_{valid} \subseteq \mathcal{O}$ will define the set of allowed operators.

\begin{algorithm}
\caption{\textbf{Greedy Hill-Climbing Algorithm for SLGBN}}
\label{alg:ghc_tabu}
\begin{algorithmic}[1]
\Require Dataset $\mathcal{D}$, initial graph $\mathcal{G}_0$, operator set $\mathcal{O}$, score $S$, patience $\eta$, size of tabu list $t$, number of B-spline bases $l$, penalization parameter $\lambda$, max. number of parents $d_{max}$.
\State $\mathcal{G}_{best} \gets \mathcal{G}_0$
\State $\mathcal{G}_{current} \gets \mathcal{G}_0$
\State $\mathcal{T} \gets \emptyset$
\State $i \gets 0$

\While{$i < \eta$}

    \State $\mathcal{O}_{valid} \gets \{\, o \in \mathcal{O}_{\mathcal{G}_{current}} : o \notin \mathcal{T},\; \max_i |\mathrm{d}_{\mathcal{G}_{current}^{o}}(X_i)| \le d_{\max},\;\mathcal{G}_{current}^{o} \text{ acyclic}  \,\}^*$
    \State $o^* \gets 
    \displaystyle
    \argmax_{o \in \mathcal{O}_{valid}}
    S\big(\mathcal{D},\, o(\mathcal{G}_{current}),\, l, \lambda\big)$
    \State $\mathcal{G}_{current} \gets o^*(\mathcal{G}_{current})$
    \If{$S(\mathcal{D}, \mathcal{G}_{current}, l, \lambda) 
        > S(\mathcal{D}, \mathcal{G}_{best}, l, \lambda)$}
        \State $\mathcal{G}_{best} \gets \mathcal{G}_{current}$
        \State $i \gets 0$
    \Else
        \State $i \gets i + 1$
    \EndIf
    \State $\mathcal{T} \gets \mathcal{T} \cup \{(o^*)^{\top}\}$

    \If{$|\mathcal{T}| > t$}
        \State remove oldest element from $\mathcal{T}$
    \EndIf
\EndWhile
\State \Return $\mathcal{G}_{best}$
\State $^{*}$ The operator $o$ is valid only if $\mathcal{G}_{current}^{\,o}$ is acyclic.
\end{algorithmic}
\end{algorithm}

The search starts from an initial structure $\mathcal{G}_0$, typically defined as the empty graph with all nodes assigned to the linear type. A predefined set of operators $\mathcal{O}$ determines the local changes that can be applied to generate the candidate structures. In hill-climbing methods, these operators usually include actions like the addition, removal or reversal of edges. In the context of Semi-Linear Gaussian Bayesian Networks (SLGBN), this framework is extended by allowing the algorithm to switch between linear and nonlinear nodes. Following~\cite{ATIENZA2022564}, we include a \texttt{TYPE\_CHANGE} operator alongside the classical \texttt{ADD}, \texttt{REMOVE}, and \texttt{REVERSE} operators.

\begin{figure}[ht!]
  \centering
  \includegraphics[scale=0.85]{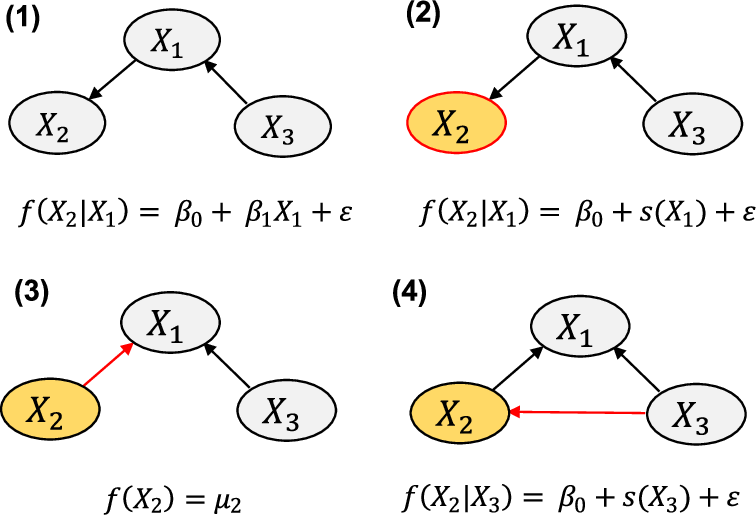}
  \caption{\textbf{Greedy Hill-Climbing:} The figure shows four steps of the Greedy Hill-Climbing Algorithm and changes in the estimation of $X_2$. Yellow nodes represent nonlinear nodes and gray nodes represent linear nodes. Changes in arcs are represented with red arrows and changes in types are represented by red contours. Step 1 starts with a linear $X_2$ node depending on $X_1$. In step 2 node type is changed from linear to nonlinear and $X_2$ is estimated using B-splines. In step 3 arc ($X_1$,$X_2$) is reversed and $X_2$ loses its only parent. Its density is at this point estimated with sample mean but the node type is still nonlinear. Note that at this point type\_change operator is not available for $X_2$ as it has no parents. In step 4 $X_2$ gains a new parent after adding ($X_3$,$X_2$) edge and is again nonlinearly estimated with B-splines, but now depending on $X_3$.}
  \label{fig:greedy_alg}
\end{figure}

For every candidate structure, model parameters are estimated, and the score is computed. Linear nodes are fitted using linear regression, and nonlinear nodes are estimated using GAM with penalized B-splines. The number of B-spline bases for each nonlinear node $l$, and the penalization parameter $\lambda$, are defined a priori and remain fixed during structure learning. Nodes without parents are automatically modeled using sample mean and variance, independently of their assigned type. Since their type will not affect their density, the type change operator is not available in these situations. Nodes will retain their node type until they get a new parent (see Figure~\ref{fig:greedy_alg}). 

\noindent
The algorithm will navigate over the candidate graph structures until the score can no longer be improved. In order to escape from local maxima, a patience parameter $\eta$ is introduced. This parameter will determine the number of consecutive iterations during which the algorithm is allowed to accept lower-scoring structures before early-stopping. In this period, the best-scoring structure is saved in cache and returned when patience limit is exceeded. This mechanism may lead to cyclic behavior, in which an operator and its inverse are repeatedly applied. To prevent this, a tabu list $\mathcal{T}$ is maintained: the last $t$ movements are stored and temporarily forbidden, discouraging immediate reversals and promoting exploration of the search space.

Despite these mechanisms, the proposed procedure remains a greedy local search strategy over the space of DAG. As a consequence, even if the underlying structure (or its equivalence class) is theoretically identifiable under a consistent scoring function, the algorithm is not guaranteed to recover it in practice, since it only explores a restricted neighborhood of the search space at each iteration. Therefore, the final solution may correspond to a local optimum of the scoring landscape rather than the globally optimal structure.

\subsubsection{Asymptotic Time Complexity}

\noindent
In this section we analyze the computational complexity of the Hill Climbing algorithm for SLGBN and we compare it with SPBN and LGBN. The asymptotic time complexity of the algorithm depends basically on two steps: (i) the event generation and (ii) the event evaluation. 

At each iteration, the algorithm considers candidate modifications to the network structure. In SLGBN, with all operators allowed and $p$ nodes, up to $p^{2}$ operations for each edge operator and $p$ node type changes are evaluated. This reduces the asymptotic complexity of event generation step to $O(p^{2})$. 

Each event requires the parameter estimation and the calculation of the score. The complexity of parameter estimation depends on the node type. In nodes with no parents, the parameters are simply estimated as sample mean and variance. With $n$ samples in the training data, the complexity is defined as $\mathcal{O}(n)$. For linear nodes, the parameters are estimated through an OLS multiple linear regression. In this case, the complexity increases quadratically depending on the number of parents $\mathcal{O}(nd^{2})$~\cite{ATIENZA2022564}. The complexity of estimating nonlinear nodes depends on two steps, the spline construction and the model fitting. Being $d$ the number of parents and $l$ the number of spline bases, the former requires $\mathcal{O}(ndl)$, while the latter requires $\mathcal{O}(n(dl)^{2})$, which dominates. Therefore, in the worst case where all nodes are estimated as nonlinear nodes, parameter estimation scales as $\mathcal{O}(n(dl)^{2})$. With $p$ nodes, this should be multiplied by $p$, $\mathcal{O}(pn(dl)^{2})$. In practice, for decomposable scores such as BIC, in each event only affected nodes are recalculated, which means recomputing only 1 or 2 nodes, so the $p$ term can be removed. The local term of EBIC* corresponds to the BIC score and is therefore decomposable, with only the affected nodes being updated. The global term, which depends on $|E|$, is updated in constant time without affecting the overall complexity. Therefore, the score evaluation per event can be performed in $O(n)$, without changing the overall complexity. Combining the number of events and the per-event cost, the asymptotic complexity order of each iteration is defined as $\mathcal{O}(p^{2}n(dl)^{2})$. For $K$ iterations, the total complexity is $\mathcal{O}(Kp^{2}n(dl)^{2})$. In practice, as $d$ and $l$ are bounded, this simplifies to:

\[
\textbf{SLGBN}: \mathcal{O}(Kp^{2}n).
\]

\noindent
In the linear Gaussian setting, node-type changes are not considered, and only $p^{2}$ structural events are evaluated. Parameter estimation is performed via OLS, with cost $\mathcal{O}(nd^{2})$ per node. As only affected nodes are re-estimated and $d$ is bounded in practice, the per-event complexity reduces to $O(n)$. The score remains local and does not alter the asymptotic cost. Therefore, the total complexity is:

\[
\textbf{LGBN}: \mathcal{O}(Kp^{2}n).
\]

\noindent
For SPBN, nonlinear nodes are estimated using kernel density estimation (KDE). The event generation process is analogous to SLGBN (except for specific nodes with forbidden type change in SLGBN), yielding an asymptotic complexity of $\mathcal{O}(p^{2})$ in selecting candidate events per iteration. The cost of KDE-based parameter estimation is $\mathcal{O}(n^{2}d^{2})$~\cite{ATIENZA2022564}. Assuming bounded $d$, this simplifies to $\mathcal{O}(n^{2})$. Thus, the total complexity is:

\[
\textbf{SPBN}:\mathcal{O}(Kp^{2}n^{2}).
\]

\noindent
Both SLGBN and LGBN scale linearly with the number of samples $n$, whereas SPBN exhibit quadratic scaling, making them more computationally expensive for large datasets. Additionally, the rate of convergence of nonparametric approaches like kernel density estimation is much slower than parametric settings and typically require sufficiently large sample sizes to provide reliable estimates~\cite{scott1992multivariate}.

\section{Methods}
\label{sec:methods}

Our goal is to evaluate the proposed Semi-Linear Gaussian Bayesian Network (SLGBN) model and assess its ability to represent data accurately, identify nonlinear node relationships, and recover the underlying graph structure. To this end, we compare SLGBN against alternative Bayesian network models. In this section, we first introduce the synthetic datasets employed in the analysis, followed by the considered models. We then describe the structure learning procedure, including the optimization algorithm and its configurations, as well as the evaluation metrics used to assess both predictive performance and structural recovery. The complete implementation of all experiments is publicly available at \url{https://github.com/digital-medicine-research-group-UNAV/SLGBN}, ensuring full reproducibility of the results.

\subsection{Data}

\noindent
To evaluate the ability of the SLGBN model to fit nonlinear Gaussian data, we implemented it on networks of varying structure and complexity. We considered three synthetic networks. While the structure of the first two networks (B5 \& B8) were manually developed for this paper, the third network (B\_insu) is the insurance network extracted from the \textit{bnlearn} network repository~\cite{bnlearn}. Table~\ref{tab:net_charac} shows the main characteristics of the three networks. The structure and parametrization of the first two networks can be found in Figure~\ref{fig:net1} and Figure~\ref{fig:net2}. The third network can be found in Figure~\ref{fig:net3}, in the appendix. While B\_insu is thought to be parametrized using discrete distributions, we developed an alternative continuous parametrization, in which linear Gaussian and nonlinear Gaussian relationships are combined additively. This parametrization can also be found in the appendix (Table~\ref{tab:Param_net3}). Obviously, the original semantic meaning of the variables is no longer preserved, as the parametrization was designed ad hoc for experimental purposes. Nevertheless, this is not relevant for our study, since our goal is to assess whether the model is able to correctly recover the underlying network structure, rather than to maintain the original interpretation of the variables. 

We generated different samples from these networks and used the resulting data to recover the original network structure. The underlying hypothesis is that, if the SLGBN model works correctly, it should achieve a good fit to the observed data and, in favorable cases, recover a structure that is close to the true underlying graph. Four training sample sizes were considered for each network (200, 1000, 5000 and 10000), and 20 independent training datasets were generated for each sample size using different random seeds. Structure learning was then performed using different models and scoring functions, and the resulting networks were evaluated on an independent validation set $\left(\mathcal{D}_\text{val}\right)$ of 1000 samples.

\begin{table}[t]
\centering
\caption{
{\bf Synthetic Networks.}}
\begin{tabular}{ cccc } 
\hline
\textbf{Network} & \textbf{Number of nodes} & \textbf{Number of edges} & \textbf{Source}\\
\hline
B5 & 5 & 5 & Own elaboration\\
B8 & 8 & 12 & Own elaboration\\
B\_Insu & 27 & 52 & bnlearn\\
\hline
\end{tabular}
\begin{flushleft} \textbf{Table notes}. Characteristics of synthetic networks.
\end{flushleft}
\label{tab:net_charac}
\end{table}

\begin{figure}[ht!]
\centering
  \includegraphics[scale=0.5]{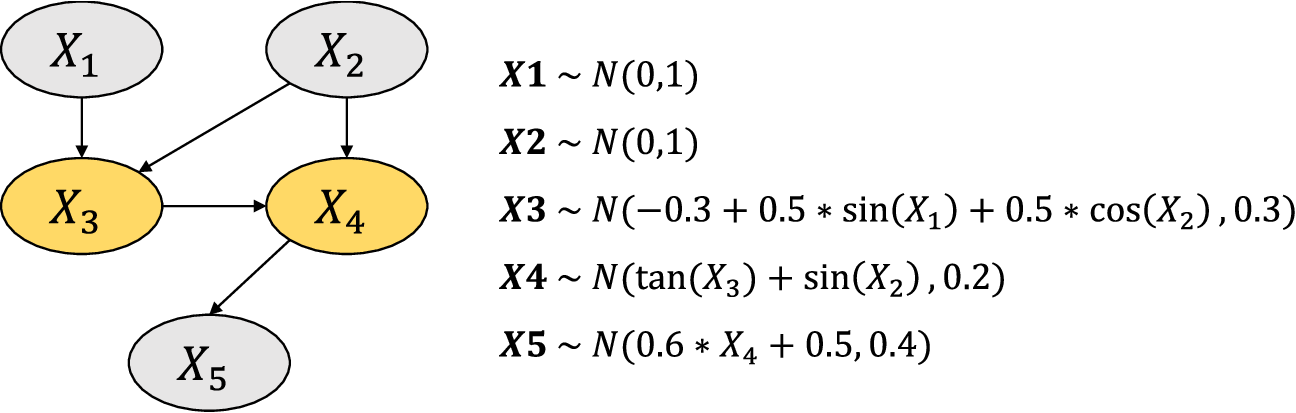}
  \caption{\textbf{B5 network.} First synthetic network structure and node probability distributions.}
  \label{fig:net1}
\end{figure}

\begin{figure}[ht!]
\centering
  \includegraphics[scale=0.4]{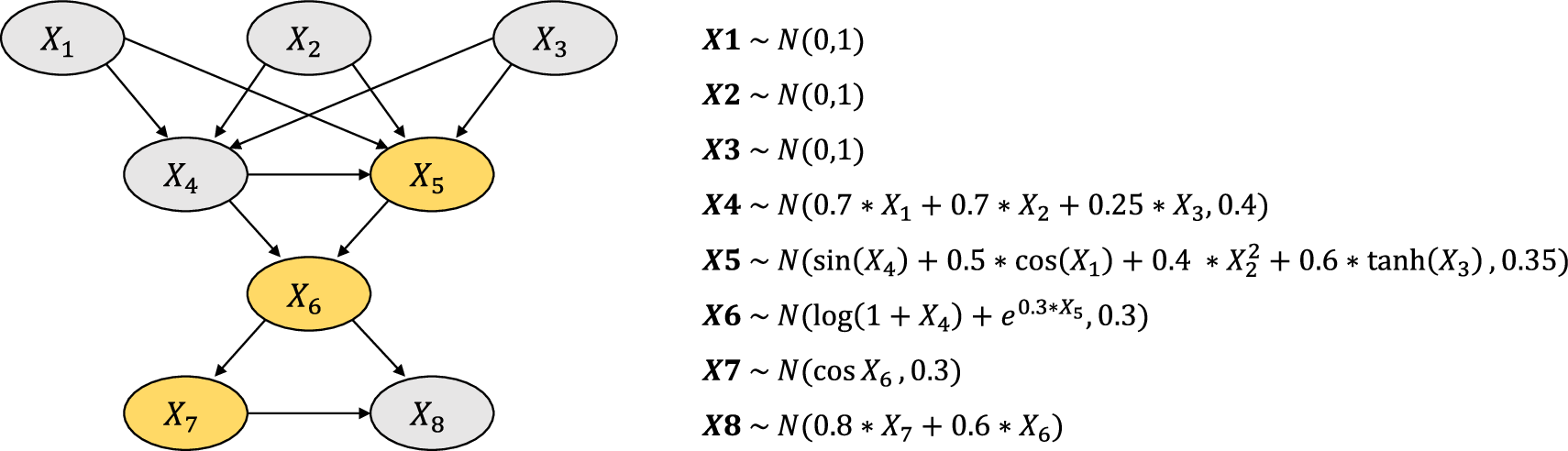}
  \caption{\textbf{B8 network.} Second synthetic network structure and node probability distribution.}
  \label{fig:net2}
\end{figure}

\subsection{Models}

We tested three different models: Linear Gaussian Bayesian Networks (LGBN), Semiparametric Bayesian Networks (SPBN) and Semi-Linear Gaussian Bayesian Networks (SLGBN). In LGBN and SLGBN, structure learning was guided independently by the BIC and EBIC* scores introduced in Section~\ref{subsec:str_lrn}. For the EBIC*, the sparsity parameter was defined at $\gamma = 0.75$. Since SPBN relies on kernel-based conditional distributions, standard information criteria such as BIC or EBIC* cannot be directly applied for structure learning; instead, the learning procedure was guided by cross-validated log-likelihood. Overall, this resulted in the evaluation of five learning configurations: LGBN-BIC, LGBN-EBIC*, SPBN-CVL, SLGBN-BIC and SLGBN-EBIC*. 

\subsection{Algorithm}

For SLGBN, we applied two variants of the Greedy Hill-Climbing (GHC) algorithm introduced in Section~\ref{subsec:str_lrn}. In the first variant, we initialize the search from the true network structure and restrict the algorithm to the \texttt{TYPE}\_\texttt{CHANGE} operator, preventing edge modifications. This setup allowed us to evaluate whether nonlinear nodes can be correctly identified when the underlying network structure is known. In the second variant, we started from an empty network and executed GHC allowing all operators. This scenario evaluated the model’s ability to learn the full network structure from scratch. If SLGBN accurately represents the original relationships among variables, it is expected to achieve a better fit and a more faithful reconstruction of the true dependency structure than other models. LGBN was also executed using the SLGBN learning algorithm, forcing all nodes to be linear. In both cases, during the executions, the patience parameter ($\eta$) was defined at 5 and the size of the tabu list ($t$) was specified as 5. In SLGBN models, the probability distribution of nonlinear nodes was estimated using the pyGAM~\cite{pygam} package, with the number of bases set to 6 and a penalization parameter of $\lambda=0.5$. SPBN networks were implemented using the PyBNesian package~\cite{pybnesian}, as described in the paper~\cite{ATIENZA2022564}. Every model started with all its nodes defined as linear.

\subsection{Evaluation Metrics}

For each model-sample size combination, we report the goodness of fit measured by the log-likelihood in the validation dataset (including its difference with respect to the ground truth) and the distance between the estimated structure and the true configuration using different metrics. We evaluated the learned structures using metrics based on the skeleton of the graph, comparing the skeleton of the learned DAG with that of the true DAG. Results are averaged across the 20 seeds for each sample size.

\begin{itemize}
\item \textbf{Log-Likelihood}
\end{itemize}

The Greedy Hill-Climbing algorithm converges to a structure $\mathcal{G}$ and a set of parameters $\boldsymbol{\hat{\theta}}$. The goodness of fit of this structure to data is measured through the log-likelihood value in the validation set: $\max_{\mathcal{G}} \mathcal{LL}\left(\{\mathcal{G}, \boldsymbol{\hat{\theta}}\} :\mathcal{D}_{val}\right)$. Although structure learning is guided by different score functions depending on the model, the resulting structures are all evaluated on the validation set using the pure log-likelihood, ensuring comparability across models.

\begin{itemize}
\item \textbf{Structural Distance}
\end{itemize}

Structural differences between the learned and true graphs are quantified using a skeleton-based Structural Hamming Distance (SHD)~\cite{tsamardinos2006max}, computed on the undirected versions of the DAG. This metric measures the number of edge additions and deletions required to transform the learned skeleton into the true skeleton, ignoring edge orientation. This choice is consistent with the identifiability limitations discussed in Section~\ref{subsubsec:Identifiability}, where multiple DAG may encode the same conditional independence structure. In SLGBN and SPBN models, not only the graph structure is learned, but also the functional type associated with each node. Accordingly, the Type Hamming Distance (THD) is also calculated, defined as the number of nodes whose type is incorrectly inferred. In SLGBN, computing the THD requires special consideration for nodes without parents, which are always modeled linearly. If a node loses its parents after being labeled as nonlinear (e.g., Figure~\ref{fig:greedy_alg}-(3)), it becomes a node labeled as nonlinear but modeled linearly. Such nodes keep their nonlinear characteristics, which will be reflected if they later have parents. Therefore, when computing THD in SLGBN, nodes are evaluated based on their assigned type rather than their effective parametrization.

To further analyze different learning patterns, we compute the precision and recall of the learned structures for each network. Precision is defined as the proportion of edges in the learned network that are correct, whereas recall corresponds to the proportion of edges in the true underlying structure that are successfully recovered. Formally, precision is given by $Pr = \text{TP}/(\text{TP}+\text{FP})$, and standard recall by $Re = \text{TP}/(\text{TP}+\text{FN})$, where TP, FP, and FN denote true positives, false positives, and false negatives, respectively. 

\section{Results}
\label{sec:results}

\begin{figure}[ht!]
\centering
  \includegraphics[scale=0.3]{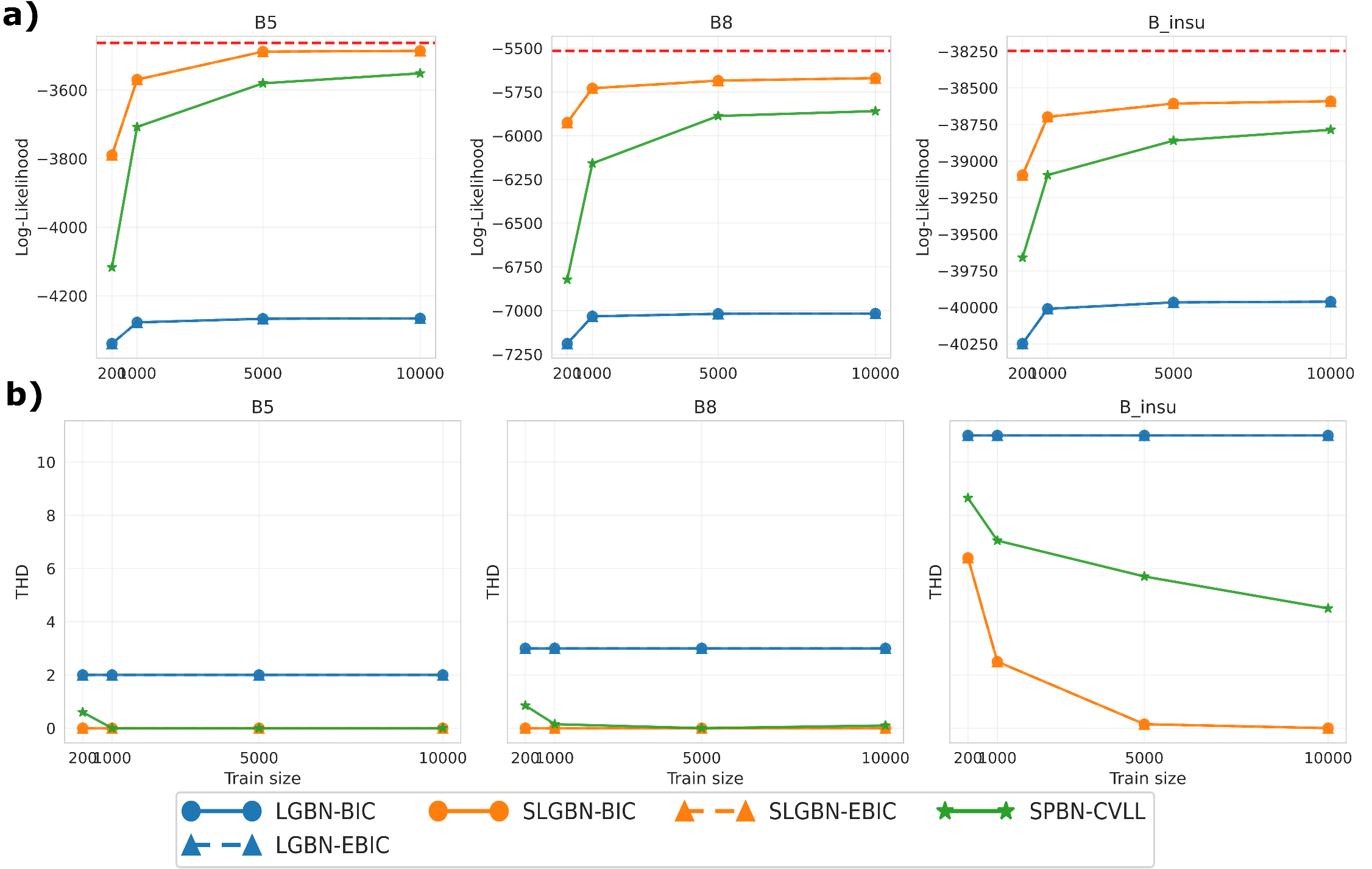}
  \caption{\textbf{Fixed metrics.} Evolution of the metrics in structure learning with fixed structure for different model-score configurations. Values represent the mean across all random seeds for each sample size. Rows correspond to metrics: a) Log-likelihood, b) THD (Lower is better). Columns correspond to networks: B5 (left), B8 (center), and B\_insu (right). X axis shows sample sizes. Y axis represents metric values, being common for THD.}
  \label{fig:fixed_metrics}
\end{figure}

\begin{figure}[ht!]
\centering
  \includegraphics[scale=0.3, trim=0 0 0 0.24cm, clip]{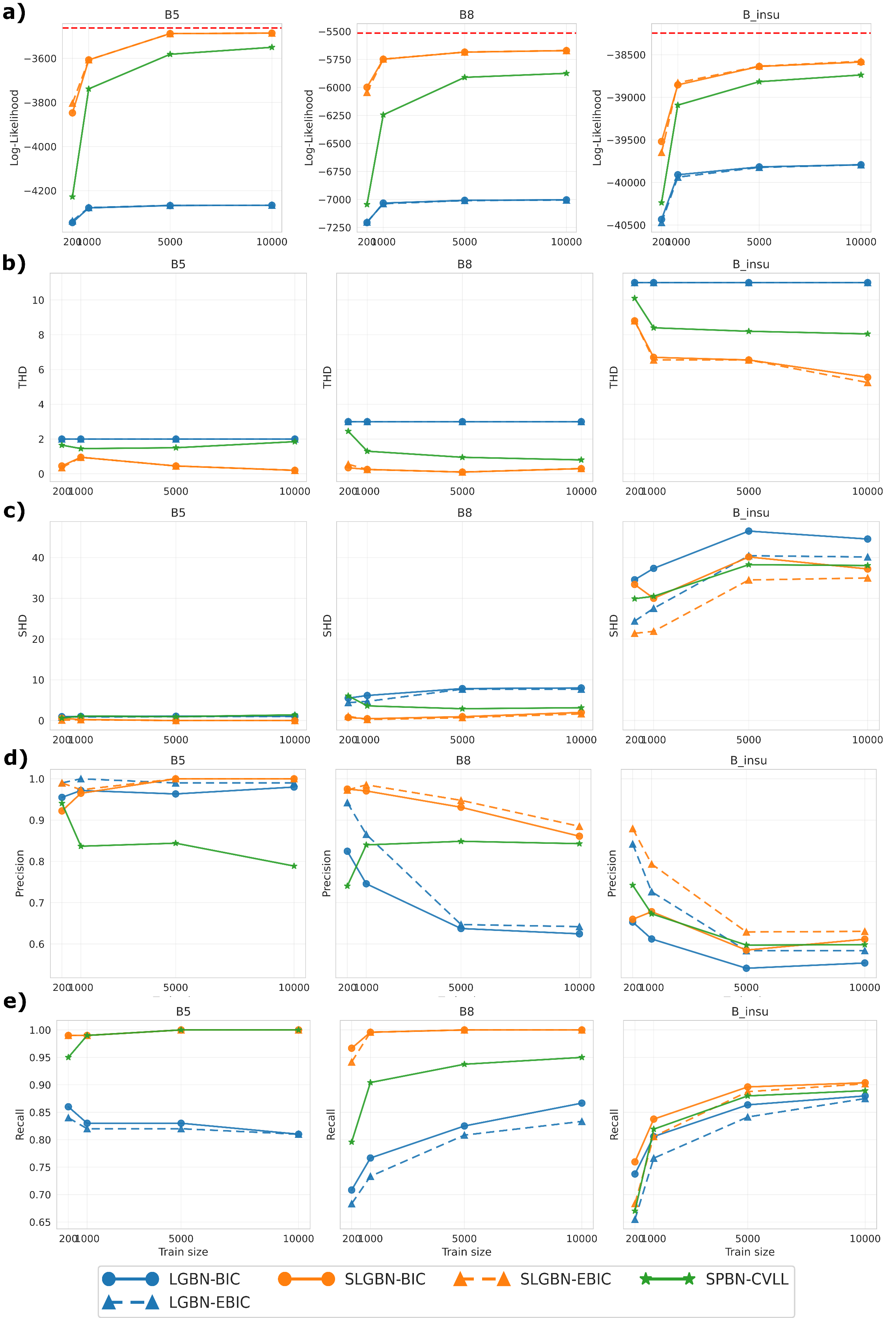}
  \caption{\textbf{Full metrics.} Evolution of metrics in full structure learning for different model-score configurations. Values represent the mean across all random seeds for each sample size. Rows correspond to metrics: a) Log-likelihood, b) THD (Lower is better), c) SHD (Lower is better), d) Precision, and e) Recall. Columns correspond to networks: B5 (left), B8 (center), and B\_insu (right). X axis shows sample sizes. Y axis represents metric values, with common axis by network for all metrics except log-likelihood.}
  \label{fig:full_metrics}
\end{figure}

This section presents the results of the learning procedures for each model in the selected evaluation metrics. All detailed numerical values can be found in the Appendix (Tables~\ref{tab:b5_fixed}–\ref{tab:b_insu_full}). Figure~\ref{fig:fixed_metrics} and Figure~\ref{fig:full_metrics} show the evolution of metrics in fixed and full approaches, respectively.

\subsection{Fixed structure learning}

For fixed-structure learning, we report the log-likelihood and the structural distance associated with node types (THD). SLGBN clearly achieves the best results, approaching the ground-truth likelihood as the sample size increases and obtaining the lowest THD values. For the two simpler networks, SLGBN correctly identifies the true types in all cases. For the more complex network, THD decreases steadily with increasing sample size, eventually converging to zero for the largest dataset. SPBN also substantially improves the LGBN model, showing similar trends but slightly worse performance than SLGBN. In LGBN, no type change is allowed, so the THD value is equal to the number of nonlinear nodes in the network. No noticeable differences are observed between the BIC and EBIC* score functions in any case, suggesting that both scores likely converge to the same structure in this setting. 

\subsection{Full structure learning}

In the full structure learning setting, we evaluate model performance using log-likelihood and structural metrics, including THD, SHD, precision, and recall. Log-likelihood follows a pattern similar to that observed under fixed-structure learning. Its value increases with sample size, progressively approaching the ground truth values. The best overall performance is consistently achieved by the SLGBN, followed by the SPBN and LGBN. Small differences emerge between the use of BIC and EBIC*, although the resulting structures in both cases still produce a very similar data fit.

Converging to the original structure is more challenging than in fixed-structure learning, as models must recover both node types and graph structure from data. In general, SLGBN provides the most consistent performance across all metrics, particularly when combined with EBIC*. A clear pattern emerges across all models: as sample size increases, recall tends to improve, while precision decreases, especially in more complex networks. This indicates that larger datasets allow models to recover more true edges, but at the cost of introducing additional false positives. This trade-off is reflected in the SHD, which does not necessarily decrease with sample size and can even increase in some settings. The EBIC* score partially mitigates this effect, achieving higher precision while maintaining competitive recall, especially in larger sample sizes. As a result, SLGBN-EBIC* consistently outperforms other models in terms of SHD.

\subsection{Discussion on results}
\label{subsec:discussion}

The observed trends show a clear divergence between log-likelihood (LL) and structural accuracy as measured by SHD. Across all settings, LL increases with sample size, which is expected since the learning procedure is explicitly driven by likelihood-based scoring functions such as cross-validated LL, BIC or EBIC*. Structural recovery, measured through SHD, is not directly optimized. Consequently, SHD does not necessarily improve with increasing sample size and, in some cases, even deteriorates. 

In the fixed structure setting, no differences are observed between BIC and EBIC* across all experiments. This behavior is expected because the graph structure is fixed and therefore the global complexity penalty introduced by EBIC* remains constant for all candidate models. As a consequence, both score functions effectively optimize the same objective and tend to converge to identical solutions.

For full structure learning, exhaustive evaluation of the DAG space is computationally infeasible, so the increasing pattern of SHD with sample size may emerge. In practice, GHC provides only a tractable local approximation to the global optimization problem and may converge to a locally optimal solution due to its restricted search neighborhood. Even under an exhaustive search over all structures, score-equivalence and other identifiability limitations (as discussed in Section~\ref{subsubsec:Identifiability}) prevent guarantees of recovering the true generating DAG from observational data alone. In addition, identifiability, when it holds, is asymptotic in nature and may not be applicable finite-sample regimes.

Within this framework, the increase in SHD can be explained by a trade-off between improved recovery of true edges and the inclusion of additional false positives as sample size increases. This effect becomes more pronounced in more complex networks, where the number of weak dependencies is higher. The global complexity penalty introduced by EBIC* helps reduce the number of spurious edges, leading to a better balance between precision and recall and therefore improving SHD in many cases. Nevertheless, this effect does not fully reverse the overall trend.

Considering these limitations, SLGBN combined with EBIC* provides the most robust performance across the considered settings, achieving the best compromise between likelihood maximization and structural accuracy. In this context, the learned models consistently achieve a strong fit to the observed data. Although recovering the true underlying DAG is not expected in general due to identifiability limitations, the estimated structures remain informative and practically useful. In particular, they capture a substantial portion of the true dependency structure and correctly identify many of the relevant conditional relationships between variables. As a result, the learned networks provide a meaningful and structured representation of the data-generating process.

\section{Conclusion}

\noindent
In this work, we introduced Semi-Linear Gaussian Bayesian Networks (SLGBN), a flexible extension of Linear Gaussian Bayesian Networks (LGBN) that maintains conditional Gaussian distributions while allowing nonlinear relationships. We also introduced a structure learning algorithm for this model and evaluated its ability to fit nonlinear Gaussian data, comparing it with other approaches such as SPBN and LGBN. Finally, we presented and discussed the results of these experiments.

The empirical results showed that SLGBN achieved strong goodness of fit and improved recovery of both node types and network structure, especially in combination with the EBIC* score. Performance was especially favorable when the underlying network structure was known, highlighting the effectiveness of the proposed estimation procedures under correct structural assumptions. However, the problem became increasingly challenging as network size grew, a limitation shared by all considered models.

From a practical perspective, the current implementation relied on hyperparameters specified a priori, such as the number of spline basis functions and the penalization parameter. While this choice ensured computational tractability, optimal performance would likely require node-specific tuning of these parameters. Such tuning could be achieved via grid search procedures but at the cost of a substantial increase in computational complexity. This approach may be feasible when the network structure is known or when dealing with small-scale networks, but it would quickly become prohibitive in larger or more complex settings. This highlights the importance of more efficient hyperparameter selection strategies to improve performance without incurring excessive computational cost.

Overall, in our experiments, SLGBN offered a promising balance between flexibility and computational efficiency, combining the advantages of both fully parametric and semiparametric Bayesian network models. However, its performance was obtained under the assumption of conditional Gaussianity and additive dependencies between parent variables, which together define the scope of the proposed framework. Our results indicate that the model behaves as expected under these assumptions, although its robustness under non-Gaussian settings or non-additive dependency structures remains an open question. Future work should investigate the behavior of SLGBN beyond these assumptions, particularly in more general distributional scenarios and real-world datasets.

\section{Funding information.} This work was supported by the Gobierno de Navarra through the ANDIA 2021 program (grant no. 0011-3947-2021-000023) and the ERA PerMed JTC2022 PORTRAIT project (grant no. 0011-2750-2022-000000).

\section{Conflicts of interest} The authors declare that they have no competing interests.

\bibliographystyle{plain}
\bibliography{bibliography.bib}

\clearpage

\begin{appendices}

\section{Additional Figures}\label{appsec1}

\begin{figure}[ht]
\centering
  \includegraphics[scale=0.75]{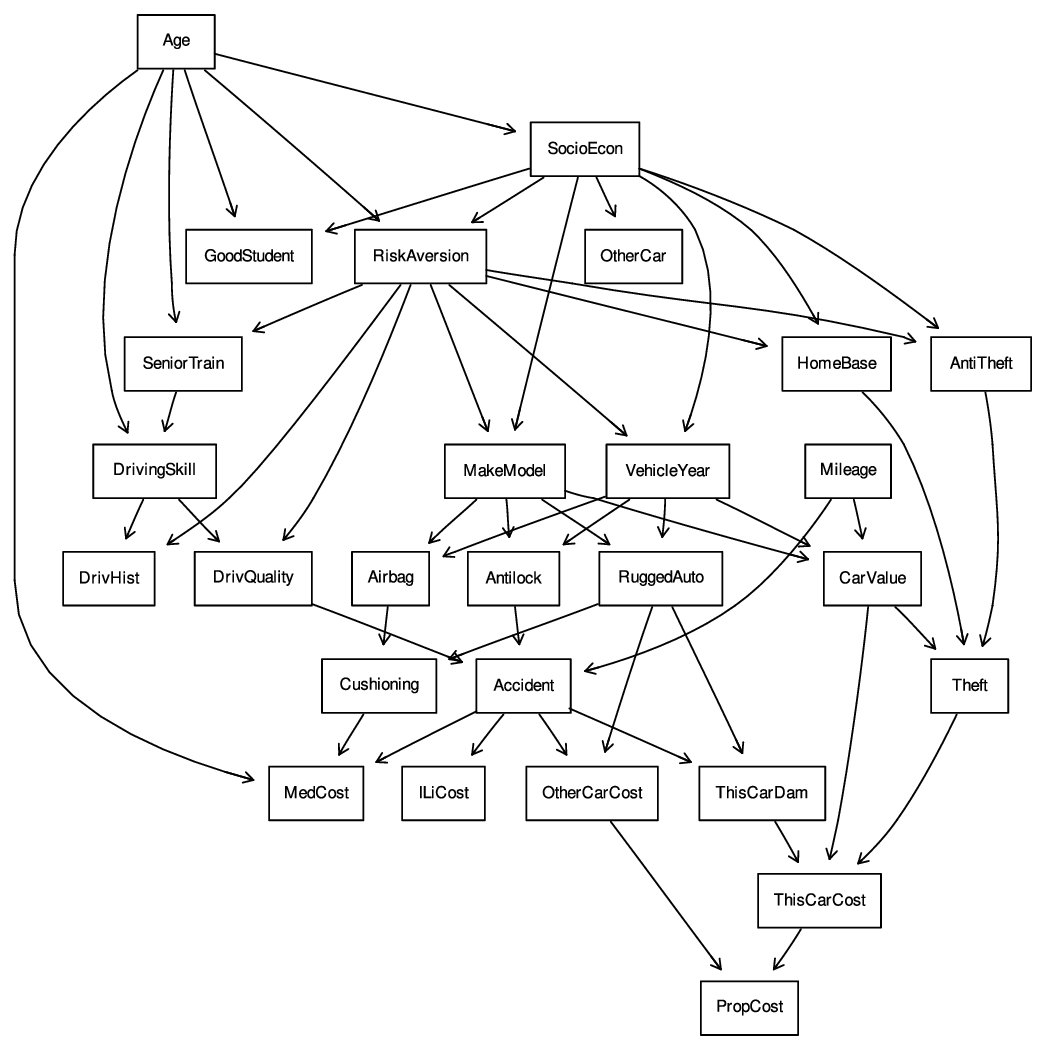}
  \caption{\textbf{BN\_insu network.} The structure of the insurance network from \textit{bnlean}. Third synthetic network.}
  \label{fig:net3}
\end{figure}

\clearpage

\section{Additional Tables}\label{appsec2}

\begin{table}[!ht]
\centering
\caption{
{\bf Parametrization of B\_insu.}}
\begin{tabular}{ ccc } 
\hline
\textbf{Node} & \textbf{Probability distribution} & \textbf{Node Type} \\
\hline
Age & $\mathcal{N}(0,1)$& Linear\\
SocioEcon & $\mathcal{N}(0.5\cdot{\sin{(Age)}} + 0.3\cdot{Age^{2}},1)$ & Nonlinear\\
RiskAversion & $\mathcal{N}(\tanh{(Age)}+0.5\cdot{SocioEcon},1)$ & Nonlinear\\
MakeModel & $\mathcal{N}(3+0.5\cdot{RiskAversion}+0.6\cdot{SocioEcon},1)$& Linear\\
Mileage & $\mathcal{N}(0,1)$& Linear\\
HomeBase & $\mathcal{N}(0.3\cdot{RiskAversion}+0.5\cdot{SocioEcon},1)$& Linear\\
SeniorTrain & $\mathcal{N}(0.3\cdot{Age} + \sin{(RiskAversion)}+0.6\cdot{SocioEcon},1)$& Nonlinear\\
VehicleYear & $\mathcal{N}(0.6\cdot{RiskAversion}+0.6\cdot{SocioEcon},1)$& Linear\\
Airbag & $\mathcal{N}(\tanh{(MakeModel)}+0.5\cdot{VehicleYear},1)$& Nonlinear\\
GoodStudent & $\mathcal{N}(0.6\cdot{Age}-0.4\cdot{SocioEcon},1)$& Linear\\
DrivingSkill &$\mathcal{N}(1+\tanh{(SeniorTrain)+0.4\cdot{Age}},1)$ & Nonlinear\\
Cushioning &$\mathcal{N}(0.8\cdot{Airbag},1)$ & Linear\\
DrivQuality & $\mathcal{N}(0.7\cdot{RiskAversion} + 0.6\cdot{DrivingSkill},1)$ & Linear\\
AntiLock &$\mathcal{N}(\sin{(VehicleYear)} + 0.5\cdot{MakeModel},1)$ & Nonlinear\\
CarValue &$\mathcal{N}(\tanh{(MakeModel)} + 0.4\cdot{Mileage} + 0.2\cdot{VehicleYear},1)$ & Nonlinear\\
AntiTheft &$\mathcal{N}\!\left(1/(1+\exp(-SocioEcon)) + RiskAversion^{2},\,1\right)$ & Nonlinear\\
Accident &$\mathcal{N}(\sin{(0.5\cdot{AntiLock} + 0.4\cdot{Mileage} + 0.4\cdot{DrivQuality})},1)$ & Nonlinear\\
Theft & $\mathcal{N}(0.6\cdot{AntiTheft} + 0.5\cdot{HomeBase} + 0.4\cdot{CarValue},1)$& Linear\\
RuggedAuto & $\mathcal{N}(0.8\cdot{VehicleYear} + 0.5\cdot{ MakeModel},1)$& Linear\\
ThisCarDam &$\mathcal{N}(0.8\cdot{Accident} + 0.3\cdot{ RuggedAuto},1)$& Linear\\
OtherCar &$\mathcal{N}(0.7\cdot{SocioEcon},1)$ & Linear\\
MedCost &$\mathcal{N}(0.6\cdot{Accident} + 0.5\cdot{Cushioning} + \sin{(Age)},1)$ & Nonlinear\\
ThisCarCost &$\mathcal{N}(0.6\cdot{ThisCarDam} + 0.5\cdot{Theft} + 0.4\cdot{CarValue},1)$ & Linear\\
OtherCarCost &$\mathcal{N}(0.8\cdot{Accident} + 0.5\cdot{RuggedAuto},1)$ & Linear\\
ILiCost & $\mathcal{N}(0.7\cdot{Accident},1)$ & Linear\\
DrivHist &$\mathcal{N}(1 + \tanh{(DrivingSkill)},1)$ & Nonlinear\\
PropCost &$\mathcal{N}(0.8\cdot{ThisCarCost} + 0.5\cdot{OtherCar},1)$ & Linear\\
\hline
\end{tabular}%
\begin{flushleft} \textbf{Table notes}. Parametrization of B\_insu network. Linearity / Nonlinearity refers to their dependence to predecessors.
\end{flushleft}
\label{tab:Param_net3}
\end{table}

\begin{table}[!ht]
\setlength{\tabcolsep}{4pt}
\centering
\caption{{\bf B5 Fixed.}}
\begin{tabular}{ccccc}
\hline
\textbf{Model-Score} & \textbf{Sample size} & \textbf{Ground LL} & \textbf{LL} & \textbf{THD} \\
\hline

LG-BIC & 200 & -3462.53 & -4339.37$\pm$61.99 & 2.00$\pm$0.00 \\
LG-BIC & 1000 & -3462.53 & -4277.49$\pm$9.75 & 2.00$\pm$0.00 \\
LG-BIC & 5000 & -3462.53 & -4266.94$\pm$2.73 & 2.00$\pm$0.00 \\
LG-BIC & 10000 & -3462.53 & -4265.84$\pm$2.58 & 2.00$\pm$0.00 \\

LG-EBIC* & 200 & -3462.53 & -4339.37$\pm$61.99 & 2.00$\pm$0.00 \\
LG-EBIC* & 1000 & -3462.53 & -4277.49$\pm$9.75 & 2.00$\pm$0.00 \\
LG-EBIC* & 5000 & -3462.53 & -4266.94$\pm$2.73 & 2.00$\pm$0.00 \\
LG-EBIC* & 10000 & -3462.53 & -4265.84$\pm$2.58 & 2.00$\pm$0.00 \\

\hline

SPBN-CVL & 200 & -3462.53 & -4116.95$\pm$207.04 & 0.60$\pm$0.68 \\
SPBN-CVL & 1000 & -3462.53 & -3707.46$\pm$45.31 & 0.00$\pm$0.00 \\
SPBN-CVL & 5000 & -3462.53 & -3580.33$\pm$12.20 & 0.00$\pm$0.00 \\
SPBN-CVL & 10000 & -3462.53 & -3551.23$\pm$11.64 & 0.00$\pm$0.00 \\

\hline

SLGBN-BIC & 200 & -3462.53 & -3789.74$\pm$71.09 & 0.00$\pm$0.00 \\
SLGBN-BIC & 1000 & -3462.53 & -3569.42$\pm$26.51 & 0.00$\pm$0.00 \\
SLGBN-BIC & 5000 & -3462.53 & -3487.87$\pm$13.71 & 0.00$\pm$0.00 \\
SLGBN-BIC & 10000 & -3462.53 & -3485.74$\pm$11.93 & 0.00$\pm$0.00 \\

SLGBN-EBIC* & 200 & -3462.53 & -3789.74$\pm$71.09 & 0.00$\pm$0.00 \\
SLGBN-EBIC* & 1000 & -3462.53 & -3569.42$\pm$26.51 & 0.00$\pm$0.00 \\
SLGBN-EBIC* & 5000 & -3462.53 & -3487.87$\pm$13.71 & 0.00$\pm$0.00 \\
SLGBN-EBIC* & 10000 & -3462.53 & -3485.74$\pm$11.93 & 0.00$\pm$0.00 \\

\hline
\end{tabular}

\begin{flushleft}
\textbf{Table notes}. Fixed structure learning results for \textbf{B5} network. Values shown as mean $\pm$ std.
\end{flushleft}

\label{tab:b5_fixed}
\end{table}

\begin{table}[!ht]
\setlength{\tabcolsep}{4pt}
\centering
\caption{{\bf B8 Fixed.}}
\begin{tabular}{ccccc}
\hline
\textbf{Model-Score} & \textbf{Sample size} & \textbf{Ground LL} & \textbf{LL} & \textbf{THD} \\
\hline

LG-BIC & 200 & -5515.31 & -7188.73$\pm$176.45 & 3.00$\pm$0.00 \\
LG-BIC & 1000 & -5515.31 & -7032.95$\pm$12.06 & 3.00$\pm$0.00 \\
LG-BIC & 5000 & -5515.31 & -7018.39$\pm$8.14 & 3.00$\pm$0.00 \\
LG-BIC & 10000 & -5515.31 & -7016.94$\pm$5.17 & 3.00$\pm$0.00 \\

LG-EBIC* & 200 & -5515.31 & -7188.73$\pm$176.45 & 3.00$\pm$0.00 \\
LG-EBIC* & 1000 & -5515.31 & -7032.95$\pm$12.06 & 3.00$\pm$0.00 \\
LG-EBIC* & 5000 & -5515.31 & -7018.39$\pm$8.14 & 3.00$\pm$0.00 \\
LG-EBIC* & 10000 & -5515.31 & -7016.94$\pm$5.17 & 3.00$\pm$0.00 \\

\hline

SPBN-CVL & 200 & -5515.31 & -6822.83$\pm$370.31 & 0.85$\pm$0.67 \\
SPBN-CVL & 1000 & -5515.31 & -6158.26$\pm$156.37 & 0.15$\pm$0.37 \\
SPBN-CVL & 5000 & -5515.31 & -5887.30$\pm$42.90 & 0.00$\pm$0.00 \\
SPBN-CVL & 10000 & -5515.31 & -5859.59$\pm$98.21 & 0.10$\pm$0.31 \\

\hline

SLGBN-BIC & 200 & -5515.31 & -5925.51$\pm$102.91 & 0.00$\pm$0.00 \\
SLGBN-BIC & 1000 & -5515.31 & -5729.16$\pm$21.06 & 0.00$\pm$0.00 \\
SLGBN-BIC & 5000 & -5515.31 & -5685.19$\pm$25.15 & 0.00$\pm$0.00 \\
SLGBN-BIC & 10000 & -5515.31 & -5671.08$\pm$27.67 & 0.00$\pm$0.00 \\

SLGBN-EBIC* & 200 & -5515.31 & -5925.51$\pm$102.91 & 0.00$\pm$0.00 \\
SLGBN-EBIC* & 1000 & -5515.31 & -5729.16$\pm$21.06 & 0.00$\pm$0.00 \\
SLGBN-EBIC* & 5000 & -5515.31 & -5685.19$\pm$25.15 & 0.00$\pm$0.00 \\
SLGBN-EBIC* & 10000 & -5515.31 & -5671.08$\pm$27.67 & 0.00$\pm$0.00 \\

\hline
\end{tabular}

\begin{flushleft}
\textbf{Table notes}. Fixed structure learning results for \textbf{B8} network. Values shown as mean $\pm$ std.
\end{flushleft}

\label{tab:b8_fixed}
\end{table}

\begin{table}[!ht]
\setlength{\tabcolsep}{4pt}
\centering
\caption{{\bf B\_insu Fixed.}}
\begin{tabular}{ccccc}
\hline
\textbf{Model-Score} & \textbf{Sample size} & \textbf{Ground LL} & \textbf{LL} & \textbf{THD} \\
\hline

LG-BIC & 200 & -38246.71 & -40247.69$\pm$58.36 & 11.00$\pm$0.00 \\
LG-BIC & 1000 & -38246.71 & -40009.88$\pm$17.00 & 11.00$\pm$0.00 \\
LG-BIC & 5000 & -38246.71 & -39965.91$\pm$4.11 & 11.00$\pm$0.00 \\
LG-BIC & 10000 & -38246.71 & -39961.14$\pm$3.62 & 11.00$\pm$0.00 \\

LG-EBIC* & 200 & -38246.71 & -40247.69$\pm$58.36 & 11.00$\pm$0.00 \\
LG-EBIC* & 1000 & -38246.71 & -40009.88$\pm$17.00 & 11.00$\pm$0.00 \\
LG-EBIC* & 5000 & -38246.71 & -39965.91$\pm$4.11 & 11.00$\pm$0.00 \\
LG-EBIC* & 10000 & -38246.71 & -39961.14$\pm$3.62 & 11.00$\pm$0.00 \\

\hline

SPBN-CVL & 200 & -38246.71 & -39660.00$\pm$73.64 & 8.65$\pm$0.99 \\
SPBN-CVL & 1000 & -38246.71 & -39095.53$\pm$32.31 & 7.05$\pm$0.60 \\
SPBN-CVL & 5000 & -38246.71 & -38860.37$\pm$10.71 & 5.70$\pm$0.57 \\
SPBN-CVL & 10000 & -38246.71 & -38785.79$\pm$12.30 & 4.50$\pm$0.69 \\

\hline

SLGBN-BIC & 200 & -38246.71 & -39096.07$\pm$64.29 & 6.40$\pm$0.75 \\
SLGBN-BIC & 1000 & -38246.71 & -38697.99$\pm$22.45 & 2.50$\pm$0.89 \\
SLGBN-BIC & 5000 & -38246.71 & -38606.37$\pm$9.84 & 0.15$\pm$0.37 \\
SLGBN-BIC & 10000 & -38246.71 & -38590.98$\pm$7.39 & 0.00$\pm$0.00 \\

SLGBN-EBIC* & 200 & -38246.71 & -39096.07$\pm$64.29 & 6.40$\pm$0.75 \\
SLGBN-EBIC* & 1000 & -38246.71 & -38697.99$\pm$22.45 & 2.50$\pm$0.89 \\
SLGBN-EBIC* & 5000 & -38246.71 & -38606.37$\pm$9.84 & 0.15$\pm$0.37 \\
SLGBN-EBIC* & 10000 & -38246.71 & -38590.98$\pm$7.39 & 0.00$\pm$0.00 \\

\hline
\end{tabular}

\begin{flushleft}
\textbf{Table notes}. Fixed structure learning results for \textbf{B\_insu} network. Values shown as mean $\pm$ std.
\end{flushleft}

\label{tab:b_insu_fixed}
\end{table}

\begin{table}[!ht]
\setlength{\tabcolsep}{3pt}
\centering
\footnotesize
\caption{{\bf B5 Free.}}
\begin{tabular}{cccccccc}
\hline
\textbf{Model-Score} & \textbf{Sample size} & \textbf{Ground LL} & \textbf{LL} & \textbf{THD} & \textbf{SHD} & \textbf{Precision} & \textbf{Recall} \\
\hline

LG-BIC & 200 & -3462.53 & -4344.83$\pm$63.74 & 2.00$\pm$0.00 & 0.95$\pm$0.76 & 0.96$\pm$0.10 & 0.86$\pm$0.09 \\
LG-BIC & 1000 & -3462.53 & -4278.18$\pm$9.59 & 2.00$\pm$0.00 & 1.00$\pm$0.46 & 0.97$\pm$0.07 & 0.83$\pm$0.07 \\
LG-BIC & 5000 & -3462.53 & -4267.58$\pm$3.54 & 2.00$\pm$0.00 & 1.05$\pm$0.39 & 0.96$\pm$0.08 & 0.83$\pm$0.07 \\
LG-BIC & 10000 & -3462.53 & -4266.25$\pm$2.73 & 2.00$\pm$0.00 & 1.05$\pm$0.39 & 0.98$\pm$0.06 & 0.81$\pm$0.04 \\

LG-EBIC* & 200 & -3462.53 & -4337.12$\pm$54.29 & 2.00$\pm$0.00 & 0.85$\pm$0.49 & 0.99$\pm$0.04 & 0.84$\pm$0.08 \\
LG-EBIC* & 1000 & -3462.53 & -4277.08$\pm$9.43 & 2.00$\pm$0.00 & 0.90$\pm$0.31 & 1.00$\pm$0.00 & 0.82$\pm$0.06 \\
LG-EBIC* & 5000 & -3462.53 & -4266.87$\pm$2.97 & 2.00$\pm$0.00 & 0.95$\pm$0.39 & 0.99$\pm$0.04 & 0.82$\pm$0.06 \\
LG-EBIC* & 10000 & -3462.53 & -4266.14$\pm$2.56 & 2.00$\pm$0.00 & 1.00$\pm$0.32 & 0.99$\pm$0.04 & 0.81$\pm$0.04 \\

\hline

SPBN-CVL & 200 & -3462.53 & -4227.83$\pm$294.20 & 1.65$\pm$0.75 & 0.60$\pm$0.94 & 0.94$\pm$0.10 & 0.95$\pm$0.14 \\
SPBN-CVL & 1000 & -3462.53 & -3738.51$\pm$61.03 & 1.45$\pm$1.00 & 1.10$\pm$0.79 & 0.84$\pm$0.10 & 0.99$\pm$0.04 \\
SPBN-CVL & 5000 & -3462.53 & -3581.58$\pm$15.46 & 1.50$\pm$0.76 & 0.95$\pm$0.39 & 0.84$\pm$0.06 & 1.00$\pm$0.00 \\
SPBN-CVL & 10000 & -3462.53 & -3550.00$\pm$16.37 & 1.85$\pm$0.37 & 1.40$\pm$0.68 & 0.79$\pm$0.07 & 1.00$\pm$0.00 \\

\hline

SLGBN-BIC & 200 & -3462.53 & -3847.44$\pm$125.88 & 0.45$\pm$0.69 & 0.55$\pm$0.83 & 0.92$\pm$0.11 & 0.99$\pm$0.04 \\
SLGBN-BIC & 1000 & -3462.53 & -3607.08$\pm$98.39 & 0.95$\pm$0.94 & 0.25$\pm$0.55 & 0.97$\pm$0.07 & 0.99$\pm$0.04 \\
SLGBN-BIC & 5000 & -3462.53 & -3487.87$\pm$13.71 & 0.45$\pm$0.60 & 0.00$\pm$0.00 & 1.00$\pm$0.00 & 1.00$\pm$0.00 \\
SLGBN-BIC & 10000 & -3462.53 & -3485.74$\pm$11.93 & 0.20$\pm$0.41 & 0.00$\pm$0.00 & 1.00$\pm$0.00 & 1.00$\pm$0.00 \\

SLGBN-EBIC* & 200 & -3462.53 & -3803.64$\pm$82.79 & 0.35$\pm$0.59 & 0.10$\pm$0.45 & 0.99$\pm$0.04 & 0.99$\pm$0.04 \\
SLGBN-EBIC* & 1000 & -3462.53 & -3606.92$\pm$98.49 & 0.95$\pm$0.94 & 0.20$\pm$0.52 & 0.97$\pm$0.07 & 0.99$\pm$0.04 \\
SLGBN-EBIC* & 5000 & -3462.53 & -3487.87$\pm$13.71 & 0.45$\pm$0.60 & 0.00$\pm$0.00 & 1.00$\pm$0.00 & 1.00$\pm$0.00 \\
SLGBN-EBIC* & 10000 & -3462.53 & -3485.74$\pm$11.93 & 0.20$\pm$0.41 & 0.00$\pm$0.00 & 1.00$\pm$0.00 & 1.00$\pm$0.00 \\

\hline
\end{tabular}
\begin{flushleft}
\textbf{Table notes}. Full structure learning results for \textbf{B5} network. Values shown as mean $\pm$ std.
\end{flushleft}

\label{tab:b5_full}
\end{table}

\begin{table}[!ht]
\setlength{\tabcolsep}{3pt}
\centering
\footnotesize
\caption{{\bf B8 Free.}}
\begin{tabular}{cccccccc}
\hline
\textbf{Model-Score} & \textbf{Sample size} & \textbf{Ground LL} & \textbf{LL} & \textbf{THD} & \textbf{SHD} & \textbf{Precision} & \textbf{Recall} \\
\hline

LG-BIC & 200 & -5515.31 & -7206.62$\pm$158.62 & 3.00$\pm$0.00 & 5.50$\pm$1.82 & 0.82$\pm$0.12 & 0.71$\pm$0.09 \\
LG-BIC & 1000 & -5515.31 & -7032.67$\pm$13.20 & 3.00$\pm$0.00 & 6.15$\pm$2.03 & 0.75$\pm$0.10 & 0.77$\pm$0.06 \\
LG-BIC & 5000 & -5515.31 & -7007.75$\pm$10.85 & 3.00$\pm$0.00 & 7.85$\pm$1.79 & 0.64$\pm$0.06 & 0.82$\pm$0.06 \\
LG-BIC & 10000 & -5515.31 & -7003.69$\pm$7.45 & 3.00$\pm$0.00 & 8.00$\pm$1.84 & 0.62$\pm$0.06 & 0.87$\pm$0.05 \\

LG-EBIC* & 200 & -5515.31 & -7201.46$\pm$165.02 & 3.00$\pm$0.00 & 4.40$\pm$1.39 & 0.94$\pm$0.10 & 0.68$\pm$0.06 \\
LG-EBIC* & 1000 & -5515.31 & -7038.44$\pm$13.25 & 3.00$\pm$0.00 & 4.70$\pm$1.69 & 0.87$\pm$0.11 & 0.73$\pm$0.06 \\
LG-EBIC* & 5000 & -5515.31 & -7009.95$\pm$10.61 & 3.00$\pm$0.00 & 7.70$\pm$1.56 & 0.65$\pm$0.06 & 0.81$\pm$0.07 \\
LG-EBIC* & 10000 & -5515.31 & -7004.81$\pm$6.96 & 3.00$\pm$0.00 & 7.70$\pm$1.66 & 0.64$\pm$0.07 & 0.83$\pm$0.05 \\

\hline

SPBN-CVL & 200 & -5515.31 & -7045.28$\pm$326.50 & 2.45$\pm$1.43 & 6.05$\pm$2.78 & 0.74$\pm$0.13 & 0.80$\pm$0.11 \\
SPBN-CVL & 1000 & -5515.31 & -6244.65$\pm$243.39 & 1.30$\pm$1.49 & 3.60$\pm$2.58 & 0.84$\pm$0.15 & 0.90$\pm$0.04 \\
SPBN-CVL & 5000 & -5515.31 & -5909.94$\pm$65.52 & 0.95$\pm$1.39 & 2.90$\pm$1.74 & 0.85$\pm$0.10 & 0.94$\pm$0.04 \\
SPBN-CVL & 10000 & -5515.31 & -5874.25$\pm$102.61 & 0.80$\pm$1.15 & 3.15$\pm$2.76 & 0.84$\pm$0.15 & 0.95$\pm$0.04 \\

\hline

SLGBN-BIC & 200 & -5515.31 & -5998.62$\pm$122.15 & 0.35$\pm$0.59 & 0.75$\pm$1.12 & 0.98$\pm$0.07 & 0.97$\pm$0.04 \\
SLGBN-BIC & 1000 & -5515.31 & -5748.06$\pm$87.65 & 0.25$\pm$0.91 & 0.45$\pm$1.00 & 0.97$\pm$0.06 & 1.00$\pm$0.02 \\
SLGBN-BIC & 5000 & -5515.31 & -5684.82$\pm$21.35 & 0.10$\pm$0.31 & 0.95$\pm$0.94 & 0.93$\pm$0.06 & 1.00$\pm$0.00 \\
SLGBN-BIC & 10000 & -5515.31 & -5671.09$\pm$21.16 & 0.30$\pm$0.47 & 2.00$\pm$0.97 & 0.86$\pm$0.06 & 1.00$\pm$0.00 \\

SLGBN-EBIC* & 200 & -5515.31 & -6044.00$\pm$145.21 & 0.55$\pm$1.19 & 1.05$\pm$1.19 & 0.97$\pm$0.06 & 0.94$\pm$0.05 \\
SLGBN-EBIC* & 1000 & -5515.31 & -5747.14$\pm$87.72 & 0.25$\pm$0.91 & 0.25$\pm$0.91 & 0.99$\pm$0.05 & 1.00$\pm$0.02 \\
SLGBN-EBIC* & 5000 & -5515.31 & -5684.30$\pm$20.48 & 0.10$\pm$0.31 & 0.70$\pm$0.73 & 0.95$\pm$0.05 & 1.00$\pm$0.00 \\
SLGBN-EBIC* & 10000 & -5515.31 & -5670.77$\pm$21.36 & 0.30$\pm$0.47 & 1.65$\pm$1.14 & 0.88$\pm$0.07 & 1.00$\pm$0.00 \\

\hline
\end{tabular}
\begin{flushleft}
\textbf{Table notes}. Full structure learning results for \textbf{B8} network. Values shown as mean $\pm$ std.
\end{flushleft}

\label{tab:b8_full}
\end{table}

\begin{table}[!ht]
\setlength{\tabcolsep}{3pt}
\centering
\footnotesize
\caption{{\bf B\_insu Free.}}
\begin{tabular}{cccccccc}
\hline
\textbf{Model-Score} & \textbf{Sample size} & \textbf{Ground LL} & \textbf{LL} & \textbf{THD} & \textbf{SHD} & \textbf{Precision} & \textbf{Recall} \\
\hline

LG-BIC & 200 & -38246.71 & -40433.97$\pm$130.31 & 11.00$\pm$0.00 & 34.55$\pm$7.79 & 0.65$\pm$0.08 & 0.74$\pm$0.04 \\
LG-BIC & 1000 & -38246.71 & -39909.71$\pm$29.50 & 11.00$\pm$0.00 & 37.35$\pm$7.11 & 0.61$\pm$0.07 & 0.81$\pm$0.02 \\
LG-BIC & 5000 & -38246.71 & -39817.52$\pm$30.12 & 11.00$\pm$0.00 & 46.50$\pm$11.14 & 0.54$\pm$0.08 & 0.86$\pm$0.02 \\
LG-BIC & 10000 & -38246.71 & -39792.27$\pm$27.43 & 11.00$\pm$0.00 & 44.55$\pm$12.21 & 0.55$\pm$0.08 & 0.88$\pm$0.02 \\

LG-EBIC* & 200 & -38246.71 & -40471.27$\pm$143.57 & 11.00$\pm$0.00 & 24.40$\pm$3.49 & 0.84$\pm$0.05 & 0.65$\pm$0.03 \\
LG-EBIC* & 1000 & -38246.71 & -39937.79$\pm$44.53 & 11.00$\pm$0.00 & 27.55$\pm$5.39 & 0.73$\pm$0.07 & 0.77$\pm$0.02 \\
LG-EBIC* & 5000 & -38246.71 & -39823.30$\pm$30.02 & 11.00$\pm$0.00 & 40.45$\pm$9.47 & 0.58$\pm$0.08 & 0.84$\pm$0.02 \\
LG-EBIC* & 10000 & -38246.71 & -39792.83$\pm$27.71 & 11.00$\pm$0.00 & 40.15$\pm$11.27 & 0.58$\pm$0.08 & 0.88$\pm$0.03 \\

\hline

SPBN-CVL & 200 & -38246.71 & -40237.64$\pm$312.16 & 10.10$\pm$1.07 & 29.90$\pm$4.94 & 0.74$\pm$0.08 & 0.67$\pm$0.05 \\
SPBN-CVL & 1000 & -38246.71 & -39091.94$\pm$50.34 & 8.40$\pm$0.68 & 30.50$\pm$4.08 & 0.67$\pm$0.05 & 0.82$\pm$0.03 \\
SPBN-CVL & 5000 & -38246.71 & -38816.69$\pm$40.35 & 8.20$\pm$1.01 & 38.25$\pm$8.88 & 0.60$\pm$0.07 & 0.88$\pm$0.03 \\
SPBN-CVL & 10000 & -38246.71 & -38737.89$\pm$43.36 & 8.05$\pm$0.94 & 38.05$\pm$10.55 & 0.60$\pm$0.07 & 0.89$\pm$0.01 \\

\hline

SLGBN-BIC & 200 & -38246.71 & -39519.32$\pm$215.21 & 8.80$\pm$1.01 & 33.40$\pm$7.50 & 0.66$\pm$0.08 & 0.76$\pm$0.04 \\
SLGBN-BIC & 1000 & -38246.71 & -38854.03$\pm$268.26 & 6.70$\pm$1.13 & 30.00$\pm$8.52 & 0.68$\pm$0.09 & 0.84$\pm$0.02 \\
SLGBN-BIC & 5000 & -38246.71 & -38637.79$\pm$75.69 & 6.55$\pm$2.14 & 40.15$\pm$12.44 & 0.59$\pm$0.10 & 0.90$\pm$0.02 \\
SLGBN-BIC & 10000 & -38246.71 & -38586.49$\pm$88.42 & 5.55$\pm$2.76 & 37.20$\pm$15.19 & 0.61$\pm$0.12 & 0.90$\pm$0.03 \\

SLGBN-EBIC* & 200 & -38246.71 & -39648.17$\pm$424.58 & 8.80$\pm$1.28 & 21.40$\pm$3.47 & 0.88$\pm$0.05 & 0.68$\pm$0.03 \\
SLGBN-EBIC* & 1000 & -38246.71 & -38828.77$\pm$213.72 & 6.55$\pm$1.15 & 21.90$\pm$8.16 & 0.79$\pm$0.11 & 0.81$\pm$0.03 \\
SLGBN-EBIC* & 5000 & -38246.71 & -38634.82$\pm$79.99 & 6.55$\pm$2.01 & 34.50$\pm$11.40 & 0.63$\pm$0.10 & 0.89$\pm$0.03 \\
SLGBN-EBIC* & 10000 & -38246.71 & -38577.58$\pm$98.36 & 5.25$\pm$2.61 & 35.00$\pm$15.68 & 0.63$\pm$0.13 & 0.90$\pm$0.04 \\

\hline
\end{tabular}
\begin{flushleft}
\textbf{Table notes}. Full structure learning results for \textbf{B\_insu} network. Values shown as mean $\pm$ std.
\end{flushleft}

\label{tab:b_insu_full}
\end{table}

\end{appendices}

\end{document}